# MCGM: MASK CONDITIONAL TEXT-TO-IMAGE GENERATIVE MODE

Rami Skaik[1], Leonardo Rossi[2], Tomaso Fontanini[3], and Andrea Prati[4]

[1234]Department of Engineering and Architecture, University of Parma, Italy
`rami.skaik`[1], `leonardo.rossi`[2], `tomaso.fontanini`[3], `andrea.prati`[4]@unipr.it

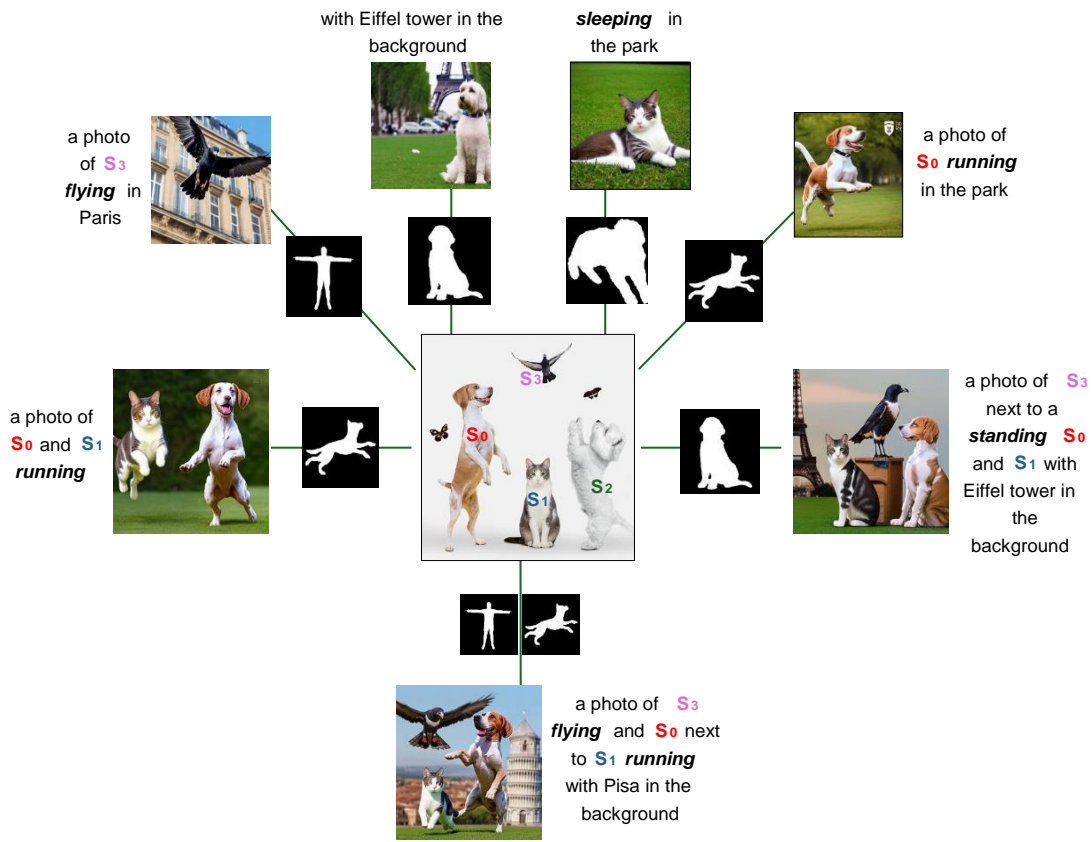

**Fig.1.** Overview of the proposed MCGM method. Starting from a single image with a set of subjects and the corresponding masks, is possible to: (a) generating any of those subjects using an arbitrary mask to set the pose, (b) generating multiple subjects at the same time using a single mask to control the pose and (c) generating multiple subjects at the same time using multiple mask to define each of their pose

## ABSTRACT

*Recent advancements in generative models have revolutionized the field of artificial intelligence, enabling the creation of highly-realistic and detailed images. In this study, we propose a novel Mask Conditional Text-to-Image Generative Model (MCGM) that leverages the power of conditional diffusion models to generate pictures with specific poses. Our model builds upon the success of the Break-a-scene [1] model in generating new scenes using a single image with multiple subjects and incorporates a mask embedding injection that allows the conditioning of the generation process. By introducing this additional level of control, MCGM offers a flexible and intuitive approach for generating specific poses for one or more subjects learned from a single image, empowering users to influence the output based on their requirements. Through extensive experimentation and evaluation, we demonstrate the effectiveness of our proposed model in generating high-quality images that meet predefined mask conditions and improving the current Break-a-scene generative model.*

## 1. INTRODUCTION

Text-to-Image generation, i.e. generating images from a text description, is a challenging yet fascinating task that has gained prominence in the field of artificial intelligence and computer vision. The ability to generate realistic images from textual descriptions has vast implications in several domains, including but not limited to, content creation, virtual reality, and design automation [2]. Over the years, researchers have developed many techniques and models to tackle this problem, each with their strengths and limitations. Conditional Text-to-Image generators have emerged as a powerful paradigm in artificial intelligence, enabling the synthesis of images based on specific textual descriptions [3,4]. In addition, there exist several other applications such as Class-Conditional Image Generation [5,6], Image Editing and Inpainting [7], Audio Synthesis [8], Text Generation [9], Climate and Weather Modeling [10], and more.

Diffusion models are a class of generative models that learn to generate data by reversing a gradual noise process. They consist of two main phases: the forward diffusion process, in which noise is gradually added to the data, and the backward process, in which the model learns to gradually denoise the data to generate new patterns. These models quickly became the state of the art of image generation due to their ability to generate high-quality and high-resolution samples and diverse patterns with a stable training process, outperforming traditional generative models such as GANs [6]. Their foundation in established theoretical principles, their flexibility in application, and their continuous improvements through ongoing research further increase their effectiveness and versatility [11]. Through training on large datasets, conditional diffusion models learn to capture complex dependencies between the conditioning information and the generated images, resulting in realistic and diverse outputs [12]. On the other side, diffusion models encounter limitations when trained on few images, as their effectiveness heavily relies on the volume and diversity of training data, which impacts their ability to generalize and produce high-quality outputs [11]. For this reason, recently, many approaches have been presented to enhance and improve diffusion models and force them to learn one or more concepts by using only few examples [13,14]. By doing so, it is possible to generate the learned concepts in different contexts using textual descriptions.

Additionally, some models can even generate and edit a specific subject by using only a single image [15]. To achieve this, these systems use transfer learning in text-to-image diffusion models, by either fine-tuning all parameters [14] or producing and optimizing a word vector [13] for a new concept. Additionally, Break-a-scene [1] can extract multiple concepts from a single image.

Nevertheless, when generating the concepts during inference, it is very difficult to specify a precise pose due to the vagueness of the text description. For example, after the diffusion model was trained to generate a specific dog, the user could want to generate a picture of that dog sitting in a very specific pose, which is very difficult to describe only through text. This limits the applicability of the aforementioned methods.

In this paper, we propose a Mask Conditional Text-to-image Generative Model (MCGM), which improves the Break-a-scene model [1] using a mask condition that specifies the target pose of the generated samples (see Fig. 1 for an overview). MCGM is able to generate a new scene starting from a single image that has one or more objects and specify the pose of these objects using the text and mask condition. By doing so, it is possible to obtain more fine-grained control of the output. More in detail, in the proposed architecture the text is responsible for the overall appearance of the scene, while the mask influences the pose of the subject. Finally, the contributions of this work are the following:

- A novel architecture capable of reproducing the subjects starting from a single image in a different context and pose using a text description and the mask of the desired pose.
- The capability of the masks to act also as weak conditioning during inference. More in detail, we can inject masks that do not strictly correspond to the desired subject, but they are still able to condition the generated samples without causing undesired artifacts.
- An additional mask cross-attention loss was introduced during training to increase the ability to link the masks with their corresponding subjects.

## 2. RELATED WORKS

### 2.1. Fine-tuning diffusion models:

Diffusion-based models [6,16–18] emerged as the new state of the art for text-to-image generation [19–22]. In this context, finetuning conditional diffusion models for text-to-image generation represents a critical area of research, aiming to enhance the performance, control, and application range of these models. Zhang et al. [23] highlighted that fine-tuning conditional diffusion models on specific datasets improves text-to-image generation performance. Then, Brown and Smith [24] provided a comparative analysis of fine-tuning strategies, offering insights into optimizing these processes, while Garcia et al. [25] emphasized the importance of hyper-parameters tuning in enhancing the performance of finetuned diffusion models. In MCGM, to generate new images of the subject in different contexts and poses, we based our work on the Break-a-scene [1] model which has two phases: the first involves designating a collection of text tokens (handles), freezing the model weights, and optimizing the handles, while in the second one, the whole model weights are fine-tuned while continuing to optimize the handles too.

### 2.2. Cross-attention

Cross-attention mechanisms enable the model to effectively align and integrate information from textual descriptions with visual data, significantly enhancing the quality of generated images from textual inputs. In the domain of text-to-image generation, cross-attention layers have been effectively utilized in various state-of-the-art models. DALL-E [26] uses cross-attention to align text descriptions with image generation, ensuring the visual output accurately reflects the input text. Then, Imagen [27] employs a cascading pipeline of cross-attention mechanisms to integrate text embeddings into the image generation process, achieving high fidelity and resolution in the generated images. Additionally, Stable Diffusion [21] utilizes cross-attention layers throughout the denoising steps, allowing the model to generate highly-detailed and semantically-accurate images from textual descriptions. Cross-attention maps in text-to-image diffusion models are used by prompt-to-prompt [20] to edit generated images. This process was then extended to real images by Mokady et al. [28]. To match text-to-image generations, attend and-excite [29] employs cross-attention maps as an explainability-based technique [30]. Furthermore, in Break-a-scene [1] cross-attention maps have the purpose of separating learned concepts. More in detail, as opposed to altering the input image, Break-a-scene concentrates on extracting textual handles from a scene and combining them into entirely new scenes. In MCGM, we utilize the cross-attention technique as in Break-a-scene, but we also incorporate the embedding of a mask image as extra information alongside text embeddings to help the model create a specific pose of the subject in the scene. Additionally, we also add a novel cross attention loss designed to better connect the masks and the corresponding subjects.

## 2.3. Personalization in diffusion models

Recently, several approaches were proposed to reproduce in different contexts and edit real images using diffusion models. In both Textual Inversion (TI) [13] and DreamBooth (DB) [14] the model learns how to generate a visual concept in different situations simply by observing images. A novel method is used by Textual Inversion [13] to embed concepts from the images into new terms in the text embedding space. By denoising supplied images, this technique increases the tokenizer dictionary's size and maximizes more tokens. Conversely, DreamBooth [14] increases the diversity of the generated results by updating the UNet parameters with a class-specific prior preservation loss and representing concepts with low-frequency words (e.g., "V"). TI and DB are simple, flexible frameworks that provide the basis for many fine-tuning-based techniques. Moreover, weight deviations during fine-tuning are examined by Custom Diffusion [31], which highlights the critical role of cross-attention layer parameters.

Furthermore, SVDiff [32] implements a successful personalization technique in the parameter space using the weight kernels' singular-value decomposition. They also give a regularization term for mixing and unmixing, which enables the generation of two concepts that are close to each other. In addition, to enable quick personalization, ELITE [15] proposed a learning-based encoder that supports input masks and encodes a single image. Its global and local mapping networks enable precise and quick custom text-to-image generation. ELITE is still unable to produce more than one concept, though. Finally, Break-a-scene [1] can extract multiple concepts from a single image using a masking mechanism. Differently from SVDiff which requires several images for each of the concepts, Break-a-scene [1] operates on a single image containing multiple concepts. MCGM follows the Break-a-scene and ELITE idea by using only a single image for the concept, in contrast to other methods that require multiple images. In addition, unlike other recent methods, the MCGM method uses a mask condition in line with the text to set the pose for one or more source subjects.

## 3. THE METHOD

### 3.1. Model Architecture

In MCGM architecture we keep the basic architecture of Break-a-scene [1] adding additional control over the pose of the generated samples. Break-a-scene goal is to extract N textual handles $\{v_i\}_{i=1}^{N}$, where the $i^{th}$ handle $v_i$ represents the concept indicated by the mask $M_i$, given a single input image I and a set of N masks $\{M_i\}_{i=1}^{N}$ indicating the subjects in the image. Once the handles are acquired, they can be utilized to direct the synthesis of images containing the concepts in novel contexts and poses during inference by inserting them into textual prompts.

In order to train MCGM to recognize multiple concepts from a single image, a small collection of image-text pairs with the format "A photo of $[v_x]$ and $[v_y]$..." is created, then the background is masked out based on the masks provided and adding a solid, arbitrary background. Each time, a random subset of tokens is chosen.

During training, the model weights $[w_i]$ and text embedding are optimized using two distinct stages. First, the model weights are frozen and the text embeddings are optimized to match the masked concepts. This enables a fast initial embedding while preserving the capability of the model, thereby offering a strong basis for the subsequent phase. Then, in the second stage, the model weights are unfrozen and optimized alongside the text tokens. This preserves editability while enabling a faithful replication of the extracted concepts. The concepts can be generated in a variety of contexts with minimal degradation in editability by using distinct text descriptions and masks.

In our work, we used the same scenario of Break-a-scene but optimized the two phases by adding a mask encoder whose purpose is encoding the mask to generate a mask embedding $[m_i]$ for each token $[v_i]$ and concatenate it with the text embedding $[t_i]$. The mask encoder projects the masks to embeddings having the same dimension as the text embeddings and then feeds both textual and mask embeddings in the cross-attention layers of the diffusion model.

By doing this, the mask embeddings will guide the model to specify the poses of the subjects. Figure 2 illustrates how we optimize the model to generate specific poses for the target subjects by combining the text and mask embeddings.

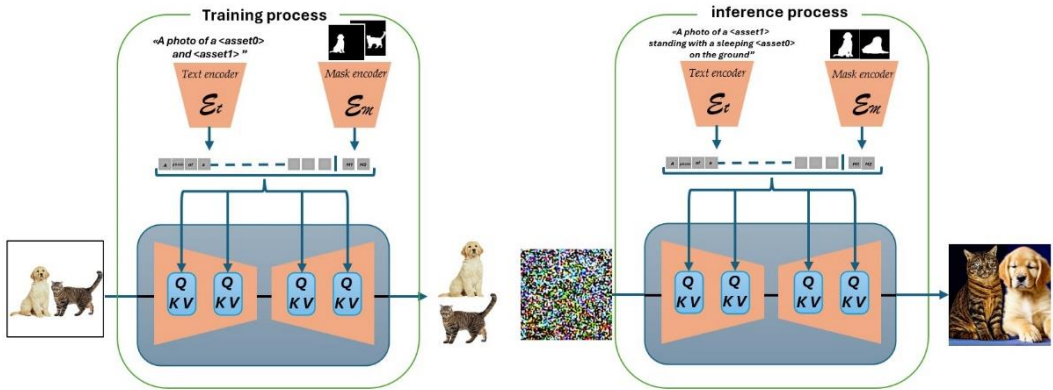

**Fig.2.** Model Architecture details: the left side is the training phase that is tasked to learn the subjects with the aid of text and mask encoders, and the right side represents the inference phase where the subjects are generated using text for expressing their context and mask to set their pose.

### 3.2. Mask Injection

In the MCGM pipeline, we added a Mask Encoder Em that aids the diffusion process by incorporating conditioning mask information into the cross-attention layers of the model. More in detail, the mask images, which are binary masks, are fed into the Mask Encoder which reduces the size of the mask and are then processed through a linear layer encoder, producing a set of masks embeddings. Finally, the text and the mask embeddings are concatenated.

The mask encoder is employed in both training and inference processes. In the training process, we generate the mask embeddings for the input concepts in the source image and use these embeddings in the two optimization phases of the model. On the other hand, in the inference process, a set of arbitrary masks is selected and embedded to specify the target pose of the generated samples and are injected alongside the text embeddings in the model. By doing so, in the inference process, the model uses the conditioning information provided by the mask encoder to generate outputs that are consistent with the mask. Loss functions described in the next sections are designed to ensure that the generated output matches the desired characteristics dictated by the mask.

## 3.3. The Optimization Losses

MCGM uses a masked version of the standard diffusion loss [33] to optimize the handles (and model weights, in the second optimization phase). This loss is calculated over the pixels of the subjects that are defined by the concepts masks:

$$L_{rec} = \mathbb{E}_{z,s,\epsilon \sim N(0,1),t}[||\epsilon \odot M_s - \epsilon_\theta(z_t, t, c_s) \odot M_s||_2^2] \quad (1)$$

where $z_t$ is the latent with added noise at time step $t$, $s$ is the number of concepts, $c_s$ is the set of embeddings obtained from text and masks, $M_s$ is the set of masks, $\epsilon$ is the noise, and $\epsilon_\theta$ is the denoising network.

Rebuilding the concepts faithfully is encouraged by using the masked diffusion loss [1] in pixel space. In addition to reconstructing the pixels of the learned subjects, the model needs to make sure that each handle only looks at the portion of the image where the related concept is located. To this end, another loss term—the Cross-Attention loss—has been introduced. This loss is defined as the MSE between the input masks and the attention calculated in the cross-attention layers of the model. The following cross-attention loss is used for each training phase:

$$L_{attn} = \mathbb{E}_{z,k,t}[||\mathcal{CA}_\theta(v_i, z_t) - M_{ik}||_2^2] \quad (2)$$

Where $\mathcal{CA}_\theta(v_i, z_t)$ is the attention map of the token $v_i$ and the noisy latent $z_t$ obtained from each cross-attention layer.

In MCGM, mask embeddings are injected with the textual embeddings for each subject in order to define its pose. This may lead to some ambiguity in the diffusion model to reconstruct each subject separately, which can occasionally result in some mixing characteristics and uncontrollable scenes, as we will explain later in section 4. Therefore, to control this process of distinguishing between the subjects in the output image, we added a mask cross-attention loss specifically designed for the mask tokens. The final cross-attention loss becomes as follows:

$$L_{Mattn} = L_{attn} + \lambda_m \mathbb{E}_{z,k,t}[||\mathcal{CA}_\theta(m_i, z_t) - M_{ik}||_2^2] \quad (3)$$

where $\lambda_m$ is the weight for the mask cross attention loss, $\mathcal{CA}_\theta(m_i, z_t)$ is the cross attention map between the mask token mi and the noisy latent $z_t$. Finally, the full training objective becomes:

$$L_{total} = L_{rec} + \lambda_{attn} L_{Mattn} \quad (4)$$

Adding $L_{Mattn}$ to the loss ensures that concept handles focus on their respective regions. We set $\lambda_{attn} = 0.01$ for accurate spatial positions in the produced image.

## 4. EXPERIMENTS

**Experimental setup**: In MCGM, we modify the official implementation of the Break-a-scene [1] model adding a mask encoder and an additional mask cross attention loss function. Concerning the input images utilized in the training and inference processes, we extracted them from different sources, some were selected from the COCO dataset [34] which contains images along with their instance segmentation masks, and some were gathered from a variety of images that included one or more subjects with suitable shapes to test different poses (such as an animal and a human). For the latter, we used the STEGO segmentation system [35] to extract the matching masks for each image in order to use them during the training or inference process. Finally, we resized each image and mask to be 512×512 pixels.

**Training details:** Regarding the learning rate, we followed the same protocol of the Break-a-scene model, detailed in Section 3, which involves two stages. Initially, in the first stage, the text embeddings are optimized with a high learning rate of $5 \times 10^{-4}$. In the subsequent stage, the UNet and text encoder weights are trained with a lower learning rate of $2 \times 10^{-6}$. Both stages use the Adam optimizer with $\beta 1 = 0.9$, $\beta 2 = 0.99$ and a weight decay of $1 \times 10^{-8}$. The number of optimization steps used is 800 steps in total, half for the first stage and half for the second stage.

### 4.1. Experiments And Discussion

**One-subject examples.** In this experiment, we used a single subject to test the effects of the mask encoder on image generation. We decided to test this scenario using different configurations:

1. *Textual prompts with coherent masks:* Fig. 3 shows several experiments using different textual prompts with an associate and coherent mask that guides the image generation. We have three subjects (dog, cat, and horse) and four pairs of prompts and masks that present different actions. In the output, we can recognize that the generated scene follows the text prompt and the subject pose is defined by the mask. Generally speaking, the model does not reproduce the injected mask shape exactly; however, it does modify the subject poses to closely resemble the mask, while allowing for an extra degree of flexibility specified by the text. On the other side, this allows to use a mask that does not represent exactly the subject, like the dog-shaped mask used for the horse in the figure, and still obtain good results.

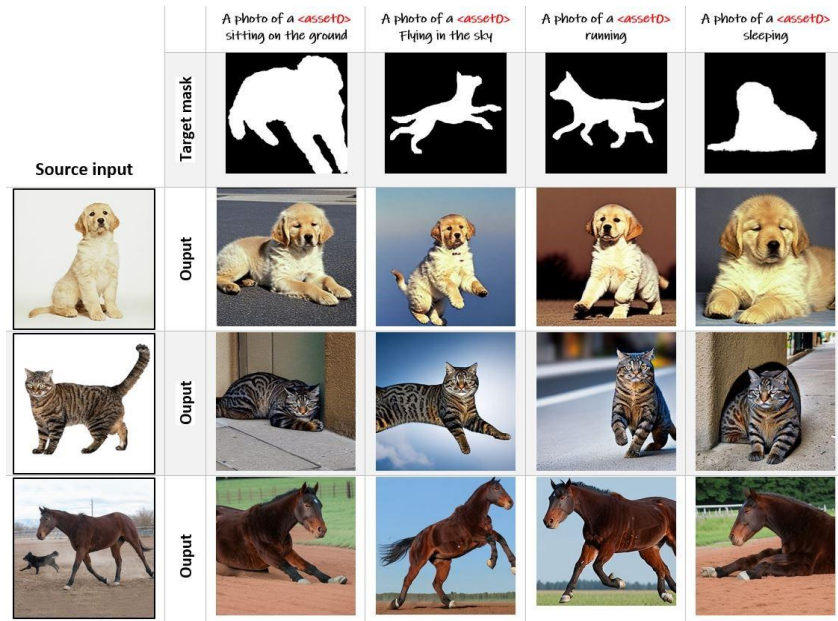

**Fig.3.** Different generated images for the input source images in different scenes based on the text prompt and with the pose guided by the target masks.

2. *Same textual prompt with different mask shapes:* Fig. 4 shows the impact of the mask encoder component in the output scene. In this case, we used the same text prompt with different mask shapes by injecting the mask image to guide the generated image to be close to the mask shape. Fig. 4(a) and 4(b) show that the dog in the input source is (a)sitting/(b)running in different positions based on the mask, although using the same text prompt.

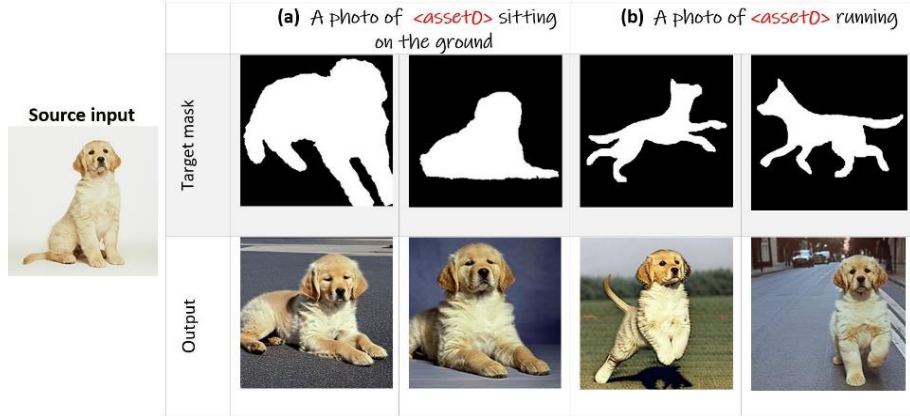

**Fig.4.** Different images generated for the same dog image using the same text prompts and different target mask poses, in (a) the dog is sitting on the ground based on the shape of two different masks. Next, in (b) the dog is running with two different poses based on the mask shape.

3. *Textual prompts with shaped-mask injection vs blank-mask/Break-a-scene:* in Fig. 5, we compare the results obtained with a normal mask describing a pose (shaped mask), the results for the MCGM model using a blank mask, and, finally, the results of Break-a-scene which does not employ a mask during inference. By doing so we verify the effect of the mask on the subject pose in the scene. The generated samples in these experiments are obtained with six pairs of prompts (three for the creature and three for the dog). It is straightforward to notice that scenes generated with the shaped masks follow the textual prompt and the mask poses, whereas, by injecting the blank masks or using the Break-a-scene model, the scene follows only the textual text by applying a random shape. Indeed, our model can be seen as a more powerful version of Break-a-scene: when using a blank mask our model is conceptually the same, but when injecting a shaped mask it can control the pose of the generated samples.

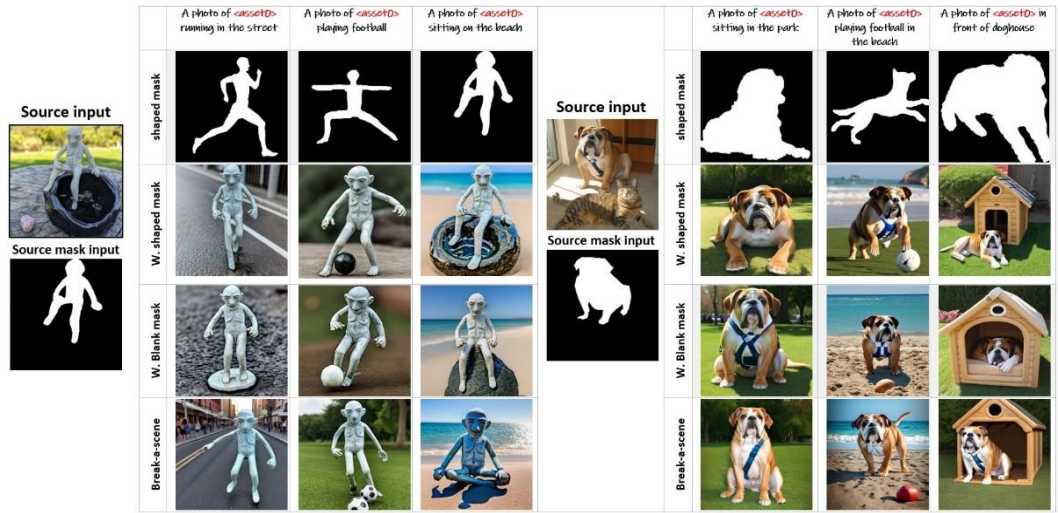

**Fig.5.** Different images generated by: MCGM with shaped mask, MCGM with Blank mask, or Break-a-scene model

4. *Textual prompt and associated mask with a general subject:* finally, the model is tested to generate images for a general subject, which means we did not use a specific subject from the source image like in previous sections, but we used only the type of subject we need (i.e., a dog), for this purpose we used different masks by using the same text prompt in each case. Fig. 6 illustrates the resulting images of the dog playing football in different positions depending on the mask injected in the generation process.

By doing so, we proved that the mask injection is crucial for determining the pose of the subject and that the model retained its full capability even after the finetuning. As stated before, the model is flexible when trying to reproduce the shape of the injected mask. For example, as shown in the right example of the figure 6, the model was asked to create an image of a dog running using a man-like pose and it was still able to generate a coherent result of a dog running following the human pose overall. As a result, in MCGM, mask injection can be seen as an additional degree of freedom added to the text prompt to create the final scene.

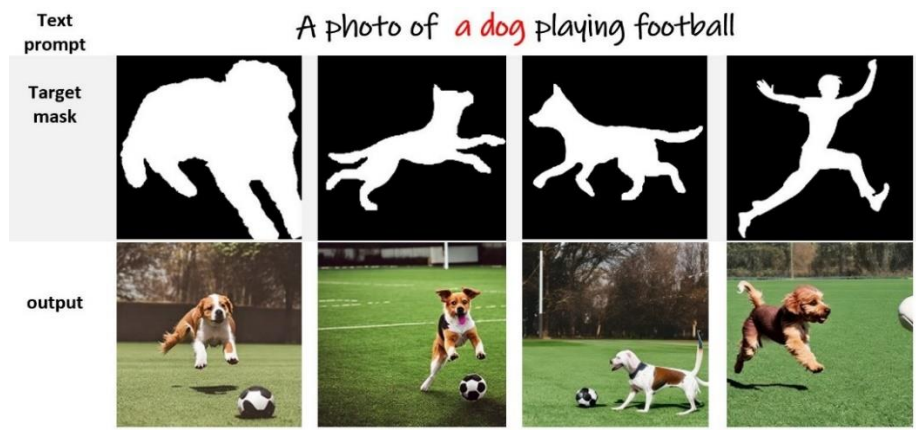

**Fig.6.** Same prompt with different target masks used to generate different positions for the dog playing football.

**Multi-subject examples**: In this section, we test the ability of the model to learn multiple subjects from a single image. In particular, two different scenarios involving multi-subject input images will be covered: (i) generating scenes with a single subject selected from the ones present in the source image and (ii) creating scenes containing multiple subjects with a different number of actions and masks used. Fig. 7 illustrates the ability to generate scenes for one subject (cat or dog) from a multi-subject source image using the mask associated with the textual prompt. Two separate source images with two different pairs of cat and dog are shown.

Fig. 8 illustrates various scenarios for generating scenes from a multi-object source image using the MCGM model. As shown in the figure, the model can create a scene for a single subject using only the text prompt without any mask, as depicted in Fig. 8(a). Additionally, it can generate a scene for a single subject by guiding the action with a mask, as shown in Fig. 8(b). The model capabilities extend beyond single-object scenes. In Fig. 8(c) and (d), the model generates scenes involving two or three different objects performing the same action, utilizing a single guidance mask.

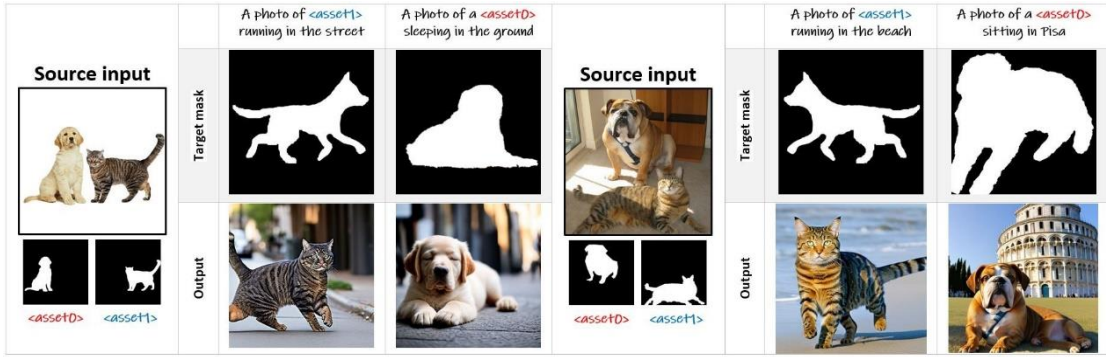

**Fig.7.** Some results for multi-subject source images used different text prompts associated with masks to generate scenes for a single subject.

Furthermore, in Fig. 8(e) and (f), the model generates scenes with two or three objects performing different actions, guided by two distinct masks.

Additionally, our model demonstrates remarkable flexibility in using a single mask shape to represent different kinds of objects. For instance, a mask designed for a dog can be used to generate a scene featuring a cat, or a human-shaped mask can be employed to create an image of a dog or bird, as illustrated in Figures 6 and 8. This adaptability highlights the model's capability to generalize across various object types, allowing for creative and versatile scene generation.

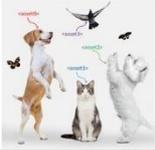

**Fig.8.** Some results for multi-subject source images used different text prompts associated with mask support the action position in the scene, each column has a different case based on the number of subjects and the number of masks used in the scene.

### 4.2. Ablation study

**Generating images with and without mask cross-attention loss in training phase**: To evaluate the effectiveness of the proposed method, in Fig. 9 we present results obtained with mask cross-attention loss, without mask cross-attention loss, and also with the baseline Break-a-scene model (that do not use the proposed mask condition injection). The top row of the figure shows three input textual prompts with a pair of masks for each prompt (one mask for each concept). The third row shows the output of the MCGM with mask cross attention loss, while the fourth row presents results without this loss, and the bottom row presents the results of the Break-a-scene model.

Indeed, without using mask cross attention loss in MCGM, the model has some problems disentangling the concepts, and the output scenes present mixed concepts as shown in Figure 9 where the cat and the dog mixed. For this reason, we added a new mask cross-attention loss during training to train the pose in addition to the cross-attention loss used in the Break-a-scene model. With this addition, in the figure, it is evident how now the model is able to disentangle the subjects perfectly following the correct pose given by the masks. On the other side, we can recognize that by using the original Break-a-scene model we can generate multiple concepts in one scene but without controlling the poses of the concepts.

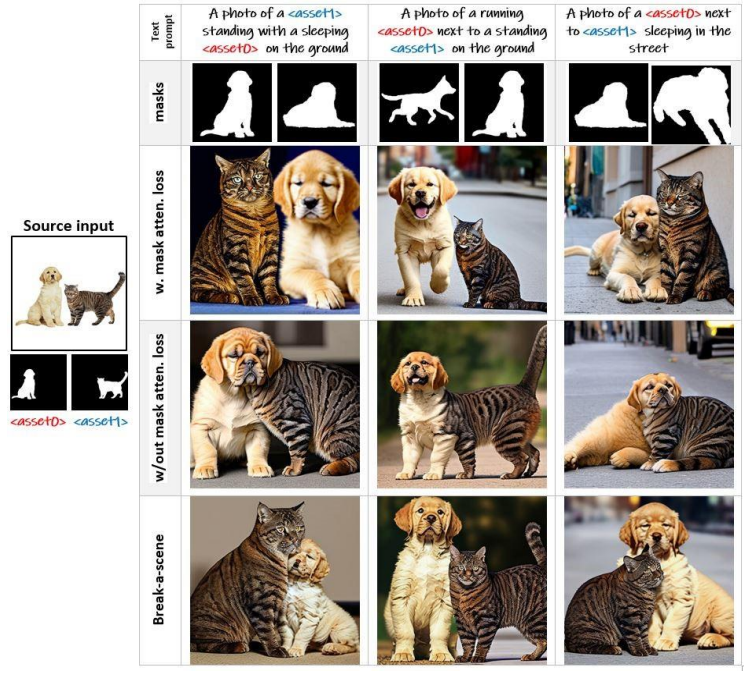

**Fig.9.** Ablation results for two-subject images using different text prompts and masks. In particular, we tested MCGM with/without mask attention loss, compared with Break-a-scene model

### 4.3. User Study

Since established and well-recognized quantitative metrics for single image text-to-image generation do not exist, we conducted a user study in order to evaluate the quality of the generated image. A random subset of our results was selected and provided to a set of anonymous users through a survey. We have envisioned two different types of questions: one multiple-choice and one based on the Likert scale [36]. The multiple-choice type has been used to evaluate the extent of the mask's effect on the generation. Given the mask and two synthesized images (MCGM vs. Break-a-scene), the evaluators were asked to select one image:" Please choose the image that is closest to the mask pose". The Likert scale (scale from 1 to 5) questions are employed to rate Text-Mask alignment: "Select the degree to which the photo result matches the input text and mask". Fifteen examples were provided, 11 for single-subject images, and 4 for two-subject images. Given the original input image, the text prompt, the mask used to generate the output, and the generated images, the evaluators were asked to choose the degree of matching between the output and the text and the mask pose used. 82 users have been involved and each of them was asked to answer 23 questions, for a total of 1886 responses.

As shown in Fig. 10, in the first part of the survey 91.3% of the evaluators chose the MCGM output image, confirming that the images generated by our method are closer to the mask shape.

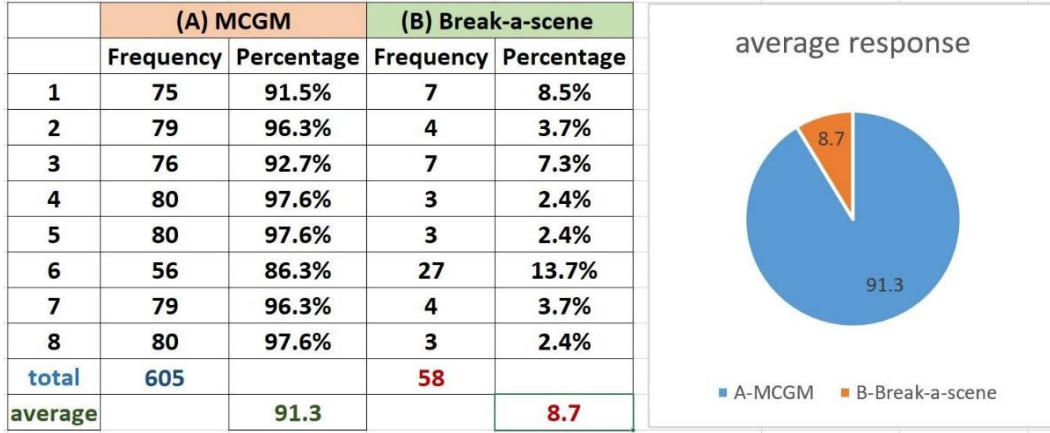

**Fig.10.** Survey results when asking which image, one generated with MCGM and the other with Break-a-scene, matches more with the pose indicated by the mask.

Next, as shown in Fig. 11, in the second part of the survey the average rating on the Likert scale is 3.8 for both one-concept and multi-concept images. These results demonstrate that MCGM generates high-quality images that correspond well to text prompts and injected masks.

| Q.no | 1 Frequency | 2 Frequency | 3 Frequency | 4 Frequency | 5 Frequency |
|---|---|---|---|---|---|
| 1 | 4 | 4 | 5 | 22 | 40 |
| 2 | 4 | 6 | 10 | 17 | 37 |
| 3 | 3 | 7 | 18 | 22 | 25 |
| 4 | 2 | 9 | 11 | 24 | 29 |
| 5 | 11 | 10 | 15 | 17 | 21 |
| 6 | 3 | 9 | 12 | 20 | 31 |
| 7 | 8 | 7 | 8 | 27 | 25 |
| 8 | 10 | 9 | 8 | 18 | 30 |
| 9 | 2 | 8 | 6 | 24 | 35 |
| 10 | 9 | 13 | 10 | 16 | 27 |
| 11 | 12 | 10 | 13 | 18 | 22 |
| one-concept | 68 | 92 | 116 | 225 | 322 |
| 12 | 2 | 7 | 17 | 25 | 24 |
| 13 | 5 | 7 | 13 | 15 | 35 |
| 14 | 5 | 15 | 13 | 19 | 23 |
| 15 | 5 | 8 | 8 | 22 | 31 |
| two-concepts | 17 | 37 | 51 | 81 | 113 |
| total | 85 | 129 | 167 | 306 | 435 |
| T-resp 1con | 823 | w-sum 1con | 3110 | Avg-rat 1con | 3.8 |
| T-resp 2con | 299 | w-sum 2con | 1133 | Avg-rat 2con | 3.8 |

**Fig.11.** Survey results when asked to determine, with a scale from 1 to 5, how much a generated sample with our method matches the pose of a target mask. *T-resp 1con*/*T-resp 2con* are the total responses for single and multiple concept images, *w-sum 1con*/ *w-sum 2con* are the weighted sum for single and multiple concept images and *Avg-rat 1con*/*Avg-rat 2con* are the average rating for single and multiple concept images.

## LIMITATION AND CONCLUSIONS

The MCGM model suffers from a directional difference issue. In some cases, the model follows the text prompt and the mask position to generate the concept in that position but struggles to generate the same orientation. As shown in Fig. 12, when using a mask condition for the subject with a side view, the model generates images for the subjects using the correct mask action position but with a front view.

Furthermore, when more than two mask conditions are used, our model fails to generate subjects in their correct position (see Fig 13. The text prompt suggests three actions for three subjects: sleeping asset1, flying asset3, and running asset0 (knowing that: asset0 represents the dog, asset1 the cat, and asset3 the bird). The mask conditions consist of three masks, one mask for each action; the results show that the positions of the mask conditions did not apply to the generated subjects.

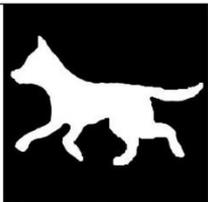

**Fig.12**. Limitation in following the correct orientation of the mask. In this case, the direction of the output images (front view) is different from the mask direction (side view) even if the overall pose is correct.

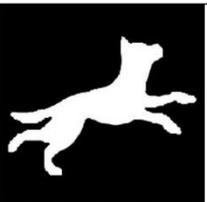

**Fig.13**. Limitation when using more than two mask conditions. Indeed, the model struggles to apply the mask conditions for all subjects in the <u>output</u> scenes.

In conclusion, the proposed MCGM model has proven to be a powerful tool for generating high-quality images with diverse mask attributes. By leveraging the conditional diffusion process and incorporating masked encoder mechanisms, this model has demonstrated a new feature compared to existing state-of-the-art generative models that use a single image for training, by adding a mask injection part to the Break-a-scene model. This feature has allowed MCGM more control and flexibility in determining the subject pose in the output images. MCGM offers a promising path for advancing the field of generative modeling and enhancing the capabilities of image synthesis and generative tasks. Further research and development in optimizing the model training procedure and scalability could potentially unlock even greater potential for this approach.

## AUTHORS

**Rami Skaik**
Ph.D. Candidate, Engineering and Architecture Department, University of Parma, Italy. He holds a Master's degree in Information Technology, earned in 2014 from the Islamic University of Gaza (IUG) in Palestine. With a robust academic background, He has been an educator since 2006, serving as a lecturer at IUG and several other universities in Palestine. his expertise lies in Multimedia Technology and AI, where he has specialized in delivering advanced coursework and fostering student success in these dynamic fields

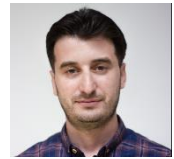

**Leonardo Rossi**
worked at CERN as Invenio core developer and collaborated with Zenodo team. He graduated in in Computer Engineering at the University of Parma in 2018. He got his PhD in Information Engineering in 2022 from the same University, working on Object Detection, Instance Segmentation techniques on social networks, funded by Adidas. He is currently a post-doc at the University of Parma, working on Anomaly Detection and Action Time Localization on video and Super-Resolution and Detection for Multi- and Hyper-spectral images.

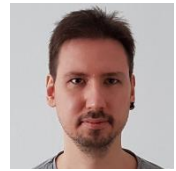

**Tomaso Fontanini**
received his Ph.D. in Information Engineering from the University of Parma in 2021 under the supervision of prof. Andrea Prati..In 2020 he was a visiting student in the VisLAB laboratory of prof. Bir Bhanu at UCR (University of California, Riverside). He is currently an Assistant Professor at the Department of Engineering and Architecture of the University of Parma. His current research interests include image generation and manipulation of facial attributes and styles in an unsupervised setting. In the past, he worked on applying meta-learning techniques to generative adversarial networks and image retrieval. He co-authored several publications about these topics in journals and international conferences. He regularly serves as a reviewer for several international conferences such as CVPR, ICVV and ECCV.

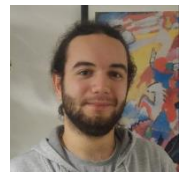

**Andrea Prati :**
graduated in Computer Engineering from the University of Modena and Reggio Emilia in 1998. He got his PhD in 2002 from the same ¡ university. He served as Assistant Professor from 2005 to 2011, and Associate Professor at the UniversityIUAV of Venice since 2015. In December 2015 he moved to the University of Parma and got promoted to full professorship in 2019. He is the head of IMPLab research group and his research interests are related to computer vision and image processing, deep learning and generative models. He is the author of 9 book chapters, 40+ papers in international referred journals and 100+ papers in proceedings of international conferences and workshops. To date, his h-index on Google Scholar is 41, with a total of 8773 citations. On Scopus, his h-index is 27, with a total of 4580 citations. Andrea Prati is Senior Member of IEEE, Fellow of IAPR, and a member of CVPL.

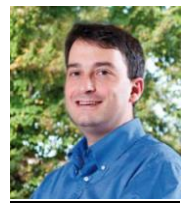